\documentclass{aims}
\usepackage{txfonts,cite}
\def\typeofarticle{Research article}
\def\currentvolume{34}
\def\currentissue{3}
\def\currentyear{2026}

\def\ppages{xxx--xxx}
\def\DOI{}
\def\Received{01 September 2025}
\def\Revised{06 December 2025}
\def\Accepted{14 February 2026}
\def\Published{}

\usepackage{graphicx}
\usepackage{caption,multirow,subcaption,graphicx}
\usepackage{amsmath,amssymb,amsfonts}
\usepackage{xcolor,hyperref}  
\usepackage{algorithm,algpseudocode}
\usepackage{mathtools, nccmath}
\usepackage{diagbox}
\usepackage{tabularx,ragged2e,booktabs}

\numberwithin{equation}{section}

\makeatletter
\renewcommand{\@biblabel}[1]{#1\hfill \hspace{-0.2cm}}
\makeatother

\usepackage{enumerate}
\usepackage{microtype}

\begin{document}

\title{Classification of systolic murmurs in heart sounds using multiresolution complex Gabor dictionary and vision transformer}

\author{%
         Mahmoud Fakhry\affil{1,}\corrauth
         and
         Abeer FathAllah Brery\affil{2}}

\shortauthors{the Author(s)}

\address{%
\addr{\affilnum{1}}{CEIEC, Universidad Francisco de Vitoria, Pozuelo de Alarcón, Madrid 28223, Spain}
\addr{\affilnum{2}}{Departamento de Informática, Universidad Carlos III de Madrid, Leganés, Madrid 28911, Spain}}

\corraddr{Email: mahmoud.fakhry@ufv.es.}

\begin{abstract}
Systolic murmurs are extra heart sounds that occur during the contraction phase of the cardiac cycle, often indicating heart abnormalities caused by turbulent blood flow. Their intensity, pitch, and quality vary, requiring precise identification for the accurate diagnosis of cardiac disorders. This study presents an automatic classification system for systolic murmurs using a feature extraction module, followed by a classification model. The feature extraction module employs complex orthogonal matching pursuit to project single or multiple murmur segments onto a redundant dictionary composed of multiresolution complex Gabor basis functions (GBFs). The resulting projection weights are split and reshaped into variable-resolution time--frequency feature matrices. Processing multiple segments of a single recording using a shared dictionary mitigates murmur variability. This is achieved by learning the weights for each segment while enforcing that they correspond to the same set of basis functions in the dictionary, promoting consistent time--frequency feature matrices. The classification model is built based on a vision transformer to process multiple input matrices of different resolutions by passing each through a convolutional neural network for patch tokenization. All embedding tokens are then concatenated to form a matrix and forwarded to an encoder layer that includes multihead attention, residual connections, and a convolutional network with a kernel size of one. This integration of multiresolution feature extraction with transformer-based feature classification enhances the accuracy and reliability of heart murmur identification. An experimental analysis of four types of systolic murmurs from the CirCor DigiScope dataset demonstrates the effectiveness of the system, achieving a classification accuracy of $95.96\%$.
\end{abstract}

\keywords{cardiovascular disease; systolic murmurs; heart sounds; multiresolution complex Gabor dictionary; vision transformer}

\maketitle
\section{Introduction}
\begin{figure}[t]
\centering
\includegraphics[width=1\textwidth,trim={0cm 0cm 0cm 0cm}]{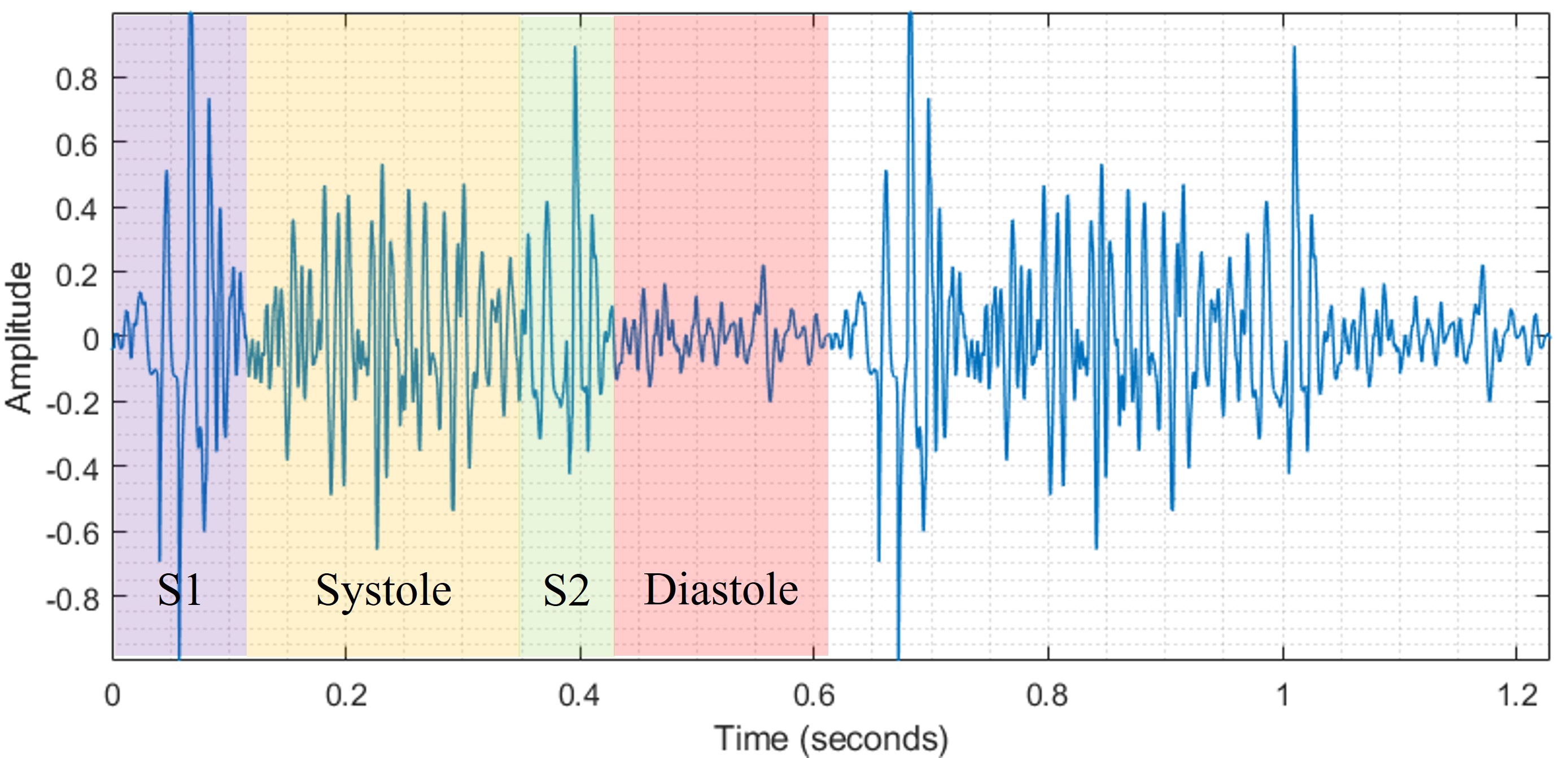}
\caption{Two cardiac cycles of heart sound, with systolic murmurs of diamond type that are located between the first ($S_1$) and second ($S_2$) heartbeats.}
\label{fig:pcg} 
\end{figure}

Cardiovascular diseases (CVDs) are among the leading causes of mortality worldwide, highlighting the urgent need for automatic diagnostic tools that enable early monitoring and detection \cite{WHO2021}. Recordings of heart sounds, or phonocardiogram (PCG) signals, offer a noninvasive and low-cost means of identifying abnormalities associated with CVDs \cite{kulkarni2010cardiac}. Heart murmurs are abnormal heart sounds caused by turbulent blood flow, which provide critical indicators of potential cardiac problems, such as valve abnormalities or congenital heart conditions. The cardiac cycle of the PCG signals consists of two primary phases: diastole (expansion) and systole (contraction), as shown in Figure \ref{fig:pcg}. Systolic heart murmurs occur during the contraction phase, between the first heartbeat ($S_1$) and second heartbeat ($S_2$), whereas diastolic heart murmurs manifest during the expansion phase, between $S_2$ and $S_1$. 

The analysis of heart sound signals has evolved significantly, progressing from the use of traditional methods such as time- and frequency-domain techniques to more sophisticated approaches in the time--frequency domain \cite{Boashash2003}. Although time-domain analysis provides valuable insight into the temporal characteristics of signals, and frequency domain methods effectively uncover spectral patterns, both approaches may face limitations in handling the nonstationary and complex nature of heart murmurs. time--frequency techniques such as the short-time Fourier transform and wavelet transform offer a more comprehensive understanding by capturing the dynamic interplay between the temporal and spectral features. These methods are particularly effective for analyzing heart murmurs that vary over cardiac cycles, thereby enhancing their characterization. 

Challenges such as individual variability and overlapping murmur components underscore the need for more advanced methods. Innovations in signal processing have provided new avenues for addressing these challenges, including dictionaries of predesigned basis functions such as Gabor functions \cite{Mallat1993MatchingPW}. By offering precise localization in both the time and frequency domains, these advanced tools enable a deeper understanding of the intricate features of heart murmurs, allowing the isolation of pathological characteristics from overlapping components \cite{Zhang1998AnalysissynthesisOT,Zhang1998TimefrequencyST}. Integrating these techniques with deep learning models enhances the accuracy of feature extraction and classification, enabling efficient automated diagnosis of heart disease, especially in resource-limited settings.

Advances in deep learning have enabled the automated analysis of signals with increasing accuracy \cite{Goodfellow}. Recurrent neural networks (RNNs) and their variants, such as long short-term memory (LSTM) networks, are used to model sequential data. However, models built on RNNs often struggle with long-term dependencies, which hinders their performance for temporally long signals. Convolutional neural networks (CNNs) have become the dominant approach for analyzing various types of signals, including heart sound signals, due to their ability to extract hierarchical features. Despite their strengths, CNNs have limitations in capturing global dependencies across the entire signal and are susceptible to overfitting, particularly when trained on small datasets. Transformer models have recently emerged as powerful tools for signal and image processing \cite{Vaswani2017AttentionIA}. Their exceptional performance is attributed to their ability to model long-range dependency and capture hierarchical representations. Transformers can capture the global context using self-attention mechanisms, thus addressing the limitations of CNNs and RNNs, which struggle with long-range dependencies. Unlike RNNs and their variants, transformers process entire sequences in parallel, enhancing their computational efficiency and making them well-suited for heart murmur analysis. A hybrid combination of CNNs for local feature extraction and transformers for global context analysis is expected to yield optimal results.

Time--frequency representations of signals are commonly employed to transform one-dimensional (1D) signals into two-dimensional (2D) image-like matrices. These structured 2D representations offer a particularly suitable input format for vision transformers (ViTs), which are deep models designed for image classification using a transformer-based architecture applied to image patches \cite{Wang2025VisionTF}. By partitioning these representations into nonoverlapping patches, ViTs can efficiently extract localized spectral and temporal features while preserving the hierarchical structure of the data. This mechanism enables the model to capture complex patterns and subtle variations in heart sound signals, thereby enhancing its ability to identify pathological conditions with high sensitivity and specificity. Additionally, the high-dimensional features derived from these time--frequency representations provide rich contextual information, further improving the robustness and interpretability of ViT-based models in analyzing heart sound signals and murmurs.

In this study, we introduce a system for the automatic classification and identification of murmurs in abnormal heart sound signals. Based on the preceding discussion, the main novel contributions of this study are as follows:
\begin{enumerate} [1)]
\item Develop a feature extraction module to calculate multiresolution time--frequency features from murmur segments using the multiresolution complex Gabor dictionary and complex orthogonal matching pursuit.
\item Develop a classification model that uses multiresolution time--frequency features to identify different types of murmurs in abnormal heart sound signals through a hybrid combination of a CNN and a vision transformer.
\item Perform a comprehensive experimental analysis of the system on a publicly available dataset of sound recordings of pathological hearts, which includes four types of systolic heart murmurs.
\end{enumerate}

{Systolic murmurs are the most common in clinical practice and are the most consistently labeled in the CirCor DigiScope dataset \cite{Oliveira2021TheCD} used in our research. They provide four well-defined morphological subclasses (crescendo, decrescendo, plateau, diamond), enabling rigorous supervised learning and benchmarking of the proposed time--frequency fusion model. In contrast, diastolic murmurs are scarce, lack consistent subclass definitions, and exhibit high intra-annotator uncertainties. For these reasons, systolic murmurs were selected as the primary target for methodological development.}

The remainder of this paper is organized as follows: Section $2$ introduces systolic heart murmurs and the dataset used to evaluate the developed system. Methods for the analysis of heart sound signals and related studies are presented in Section $3$. In Section $4$, we explain the developed systems in detail. The experimental analysis and results are discussed in Section $5$. Finally, the conclusions of this study are summarized in Section $6$.

\section{Systolic heart murmurs and dataset}
\begin{figure*}[t]
\centering
\includegraphics[scale=0.6,trim={0cm 0cm 0cm 0cm}]{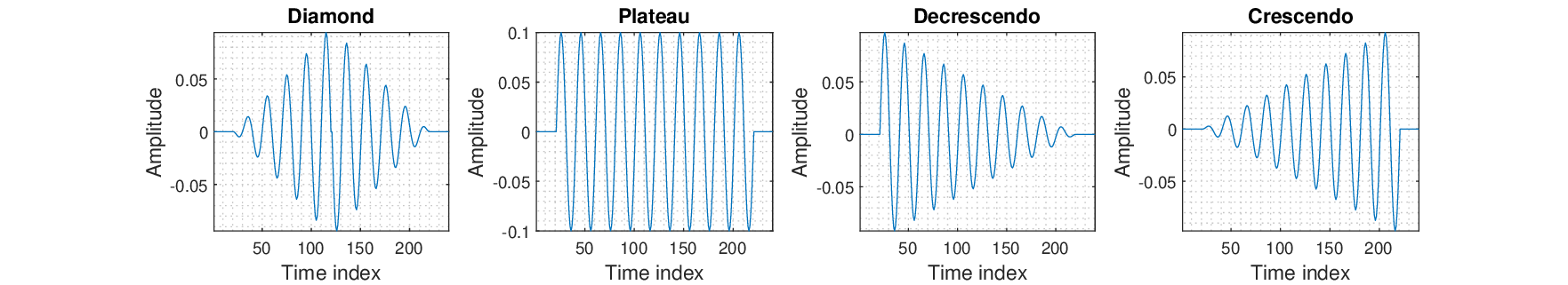}
\caption{Shapes of different types of systolic heart murmurs in the time domain.}
\label{fig:mur}
\end{figure*}

Systolic heart murmurs manifest as structural abnormalities, such as valvular stenosis or regurgitation, which disrupt normal unidirectional blood flow \cite{dd8d8}. For instance, aortic stenosis produces harsh crescendo--decrescendo murmurs as blood is forcefully ejected through a narrowed aortic valve. Simultaneously, mitral regurgitation generates holosystolic murmurs due to backflow of blood into the left atrium during ventricular contraction. In some cases, systolic murmurs may also be caused by high-output states, such as hyperthyroidism, which increases blood flow velocity, or by congenital defects such as ventricular septal defects. These sounds vary widely in intensity, pitch, and quality, ranging from soft and blowing to loud and harsh, and often require careful auscultation and diagnostic techniques for their accurate identification and assessment. Understanding the characteristics and causes of systolic murmurs is critical in clinical practice, as they can serve as important clues for diagnosing and managing heart conditions. Below is a detailed description of the four common patterns of systolic, which are represented in Figure \ref{fig:mur} as follows: 
\begin{itemize}
\item \textbf{Diamond}: These murmurs increase in intensity during the early part of systole, reach a peak, and then decrease. They are often associated with aortic or pulmonic stenosis, in which turbulent blood flow occurs because of a narrow valve opening. The characteristics of these murmurs reflect changes in blood flow velocity during systole, providing clues for stenosis identification.
\item \textbf{Plateau}: These murmurs maintain a relatively constant intensity throughout systole, resembling a flat or rectangular profile when visualized on a phonocardiogram. They are frequently associated with mitral regurgitation or ventricular septal defects, where consistent regurgitation flow or shunting occurs due to defective valve closure or septal abnormalities, respectively.
\item \textbf{Decrescendo}: These murmurs start loudly at the beginning of systole and gradually decrease in intensity as systole progresses. They are typically associated with conditions such as aortic or pulmonic regurgitation, in which blood leaks backward into the ventricles during systolic contractions. This declining intensity reduces the pressure gradient as the ventricle empties.
\item \textbf{Crescendo}: These murmurs begin softly and progressively increase in intensity, peaking either midsystole or late in systole. This pattern is associated with aortic stenosis or hypertrophic cardiomyopathy, in which the obstruction to blood flow progressively increases as the heart contracts. The crescendo shape highlights the increasing severity of turbulence during systole.
\end{itemize}
\subsection{Dataset}
The dataset used in the experimental analysis was extracted from the CirCor DigiScope dataset \cite{Oliveira2021TheCD}. Heart sound recordings were acquired from four anatomical locations on the chest: the aortic (AP), pulmonary (PP), mitral (MP), and tricuspid points (TP). These recordings were sampled at a high resolution of $4000$ Hz to capture the fine details of heart sound characteristics. They were normalized to the range [$-1$, $1$] to standardize the amplitude of the input signal. The normalized PCG signals were then automatically segmented using state-of-the-art algorithms to accurately determine the boundaries and positions of the heartbeat components $S_1$ and $S_2$ as well as the systole and diastole phases, ensuring precision in the extraction of clinically significant segments. 

All heart sound recordings were systematically screened for murmurs at each auscultation site, with the majority defined as having systolic murmurs. Although diastolic murmurs have a lower prevalence in the dataset, they are clinically critical, often signaling severe conditions such as aortic regurgitation or mitral stenosis. To evaluate the feasibility of extending our framework, we examined the annotation structure of the dataset. Although $34$ recordings ($\sim$8.9\% of murmur-positive cases) were labeled as containing \textit{diastolic murmurs}, none include morphological subtype annotations. In contrast, systolic murmurs have $4337$ expert-annotated segments across four well-defined shape classes, enabling supervised shape-based classification.

Each recording was carefully segmented to isolate multiple systolic murmur segments, which were assigned a specific label corresponding to one of the four predefined murmur types. This structured labeling process ensures consistency and accuracy in data annotation, providing a robust framework for subsequent pattern recognition and classification processes. Table \ref{tab:data} summarizes the dataset composition, detailing the total number of signal samples, distribution of systolic segments with various murmur patterns, and specific locations where they were recorded. The average segment length is approximately $500$ samples, highlighting the short but information-rich nature of these segments. 
\begin{table*}
\centering
\caption{Number of signal samples/number of segments for all types of murmurs.}
\label{tab:data}
\begin{tabular}{llllllllll}
\hline
\multirow{2}{*}{\text{Location}}&\multicolumn{4}{c}{\text{Patterns}}&\multirow{2}{*}{\text{Total}}\\
&\text{Diamond}&\text{Plateau}&\text{Decrescendo}&\text{Crescendo}&\\
\hline
\text{AP}&9/167&7/91&2/87&1/14&19/359\\
\text{PP}&10/283&36/787&14/420&1/13&61/1503\\
\text{MP}&4/32&30/750&8/293&0&42/1075\\
\text{TP}&8/164&38/979&10/257&0&56/1400\\
\text{Total}&31/646&111/2607&34/1057&2/27&178/4337\\
\hline
\end{tabular}
\end{table*}

\section{Analysis of heart sound signals and related studies} 
Traditional methods for analyzing heart sound signals include time, frequency, and time--frequency domain techniques. Time-domain analysis involves analyzing the amplitudes of the waveforms \cite{Rangayyan1990,Zheng2015ANH}. This approach can identify transient features such as the onset and offset of murmurs. Frequency-domain methods, such as the Fourier transform, reveal the spectral components of signals \cite{Stein1982FrequencyCO,Akay1996}. These methods excel in analyzing periodicity and dominant frequencies, which are often associated with structural anomalies, such as valve defects. However, they assume signal stationarity, which limits their applicability to nonstationary signals. time--frequency domain techniques, including the short-time Fourier transform and wavelet transforms, unify temporal and spectral information for the comprehensive analysis of nonstationary signals \cite{Fakhry2023VariationalMD,Atanasov2008IsolationOS}. These methods enable the dynamic monitoring of frequency components over time, allowing for a more accurate characterization of murmurs that evolve during each cardiac cycle. 

Efficient analysis of PCG signals, particularly in the presence of overlapping murmurs and other cardiac events, requires advanced techniques \cite{HaghighiMood1997TimefrequencyAO}. Recent advances in signal processing have introduced dictionaries of predesigned basis functions for signal analysis \cite{Daugman1985}. Among these, the Gabor dictionary, which comprises real Gaussian-windowed sinusoids localized in both time and frequency, is particularly suitable for analyzing heart murmurs \cite{Fakhry2022AnalysisOH,Fakhry2024ElasticNR}. The complex Gabor dictionary extends the traditional dictionary by incorporating complex-valued functions and providing additional phase information. This added dimension of analysis helps distinguish between different feature patterns in complex signal data. In \cite{mah2024}, we proposed an overcomplete multiresolution complex Gabor dictionary composed of complex Gabor functions with diverse time--frequency resolutions to create a hyperspectral space for detailed signal analysis. Using complex orthogonal matching pursuit, a systolic murmur segment is represented as a sparse, complex vector based on the dictionary. The calculated energy distribution across resolutions revealed that each type of systolic murmur exhibited unique patterns, enabling its differentiation.

In \cite{Jabbari2011ModelingOH}, the authors proposed a mathematical model for extracting crucial features from systolic murmurs using a Gabor dictionary. The extracted features were classified using a feedforward multilayer perceptron. In \cite{Shabbir2023HeartMC}, the classification of heart murmurs was thoroughly explored using selected convolutional neural network models, such as VGGNet and ResNet. This study leveraged various signal representations, including spectrograms, Mel-frequency cepstral coefficients, and short-time Fourier transforms. A multiscale attention convolutional compression network was proposed in \cite{Yin2024DetectionOC} to detect coronary artery disease. The network uses a multiscale convolution structure to capture comprehensive features and a channel attention module to enhance these features. Vision transformers (ViTs) were used to classify heart sound signals based on spectrograms in \cite{Kim2022ClassificationOP}. Similarly, ViTs have been applied to classify PCG signals using bispectrum-inspired features \cite{Liu2023HeartSC}. In \cite{Han2025ENACTHeartE}, the authors proposed an ensemble approach for classifying PCG signals. This method leverages the complementary strengths of CNNs and ViTs using a mixture of expert frameworks. A hybrid convolution--transformer neural network was introduced in \cite{Zhao2025DetectionOC}. The study demonstrated the feasibility of using a transformer-based method to detect coronary heart disease in a clinical dataset of heart sound recordings. In \cite{Wang2023PCTMFNetHS}, the authors introduced a parallel CNN transformer network to classify PCG signals. They applied a second-order spectral analysis and used a transformer along with two CNNs to extract hierarchical features, which were then fused for final classification.

Expanding on our analysis of systolic heart murmurs presented in \cite{mah2024}, we propose a novel feature extraction module that uses a multiresolution complex Gabor dictionary in conjunction with complex orthogonal matching pursuit to capture intrinsic time--frequency features with different resolutions. Furthermore, inspired by the widespread success of vision transformers in various domains, we introduce a hybrid feature classification model that integrates convolutional neural networks with a vision transformer to capture local feature patterns and enhance global contextual understanding.
\begin{figure}[t!]
\centering
\includegraphics[width=0.85\textwidth,trim={0cm 0cm 0cm 0cm}]{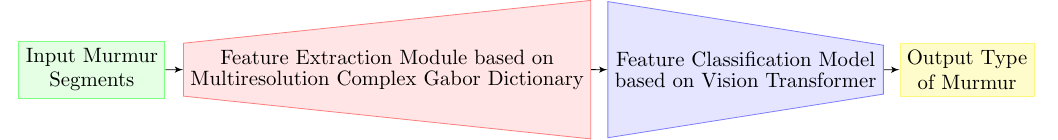}
\caption{Block diagram of the developed system.}
\label{fig:sys} 
\end{figure} 
\section{Developed system}
We present an innovative system tailored for the classification and identification of systolic heart murmurs by leveraging advanced signal processing and deep learning techniques. The system consists of two key components: a feature extraction module and a feature classification model. The feature extraction module systematically analyzes systolic heart murmurs and transforms them into representative time--frequency feature matrices using advanced decomposition and transformation techniques. These extracted features serve as inputs for the classification model, which employs deep learning models to achieve high accuracy and robustness. Each component is meticulously designed to enhance performance and ensure the reliable identification of systolic murmurs in diverse patient populations. Figure \ref{fig:sys} shows a block diagram of the proposed system.
\subsection{Feature extraction module}
\begin{figure*}[t!]
\centering
\includegraphics[width=\textwidth,trim={0cm 0cm 0cm 0cm}]{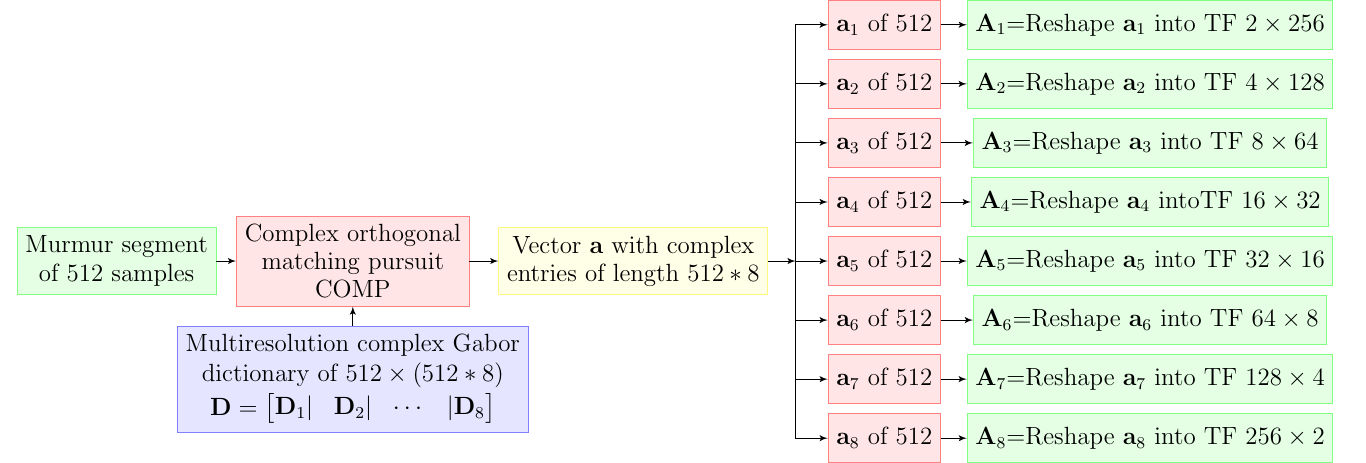}
\caption{Block diagram of the developed feature extraction module, mainly built based on complex orthogonal matching pursuit in conjunction with a multiresolution complex Gabor dictionary. The segment length is $512$ samples, resulting in a Gabor dictionary with $8$ different time--frequency resolutions.}
\label{fig:module} 
\end{figure*}

\begin{figure*}[t!]
\centering
\includegraphics[scale=0.61,trim={0cm 0cm 0cm 0cm}]{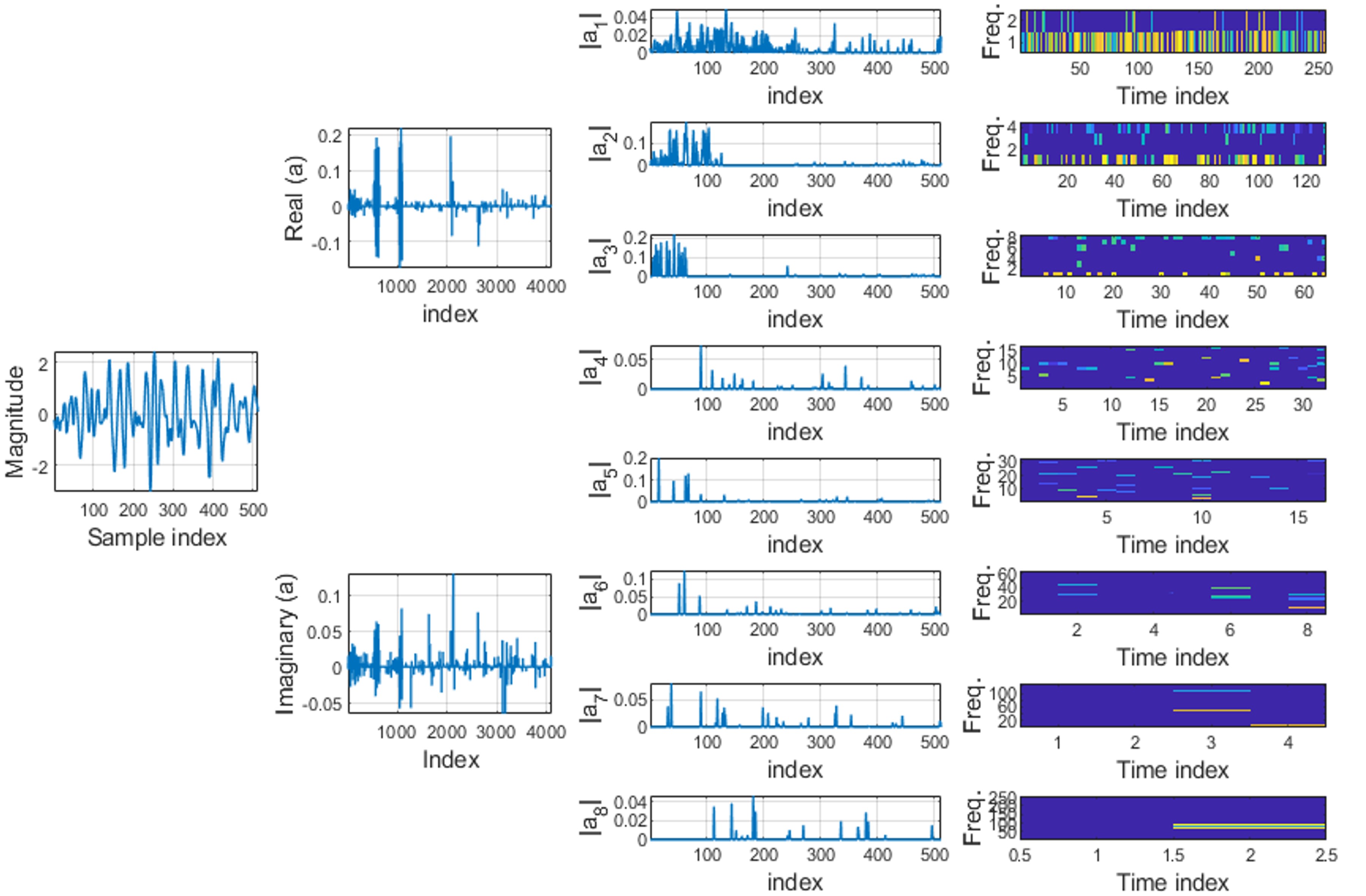}
\caption{Feature extraction from a murmur segment of diamond type using COMP and the developed multiresolution complex Gabor dictionary. Columns from left to right: murmur segment, real and imaginary parts of the vector of projection weights $\mathbf{a}$, absolute value of $\mathbf{a}_j$, and reshaped time--frequency representations $\mathbf{A}_j$ of the absolute of $\mathbf{a}_j$.}
\label{fig:segment}
\end{figure*}

\begin{figure*}[t!]
\centering
\includegraphics[scale=0.70,trim={1.9cm 0cm 0cm 0cm}]{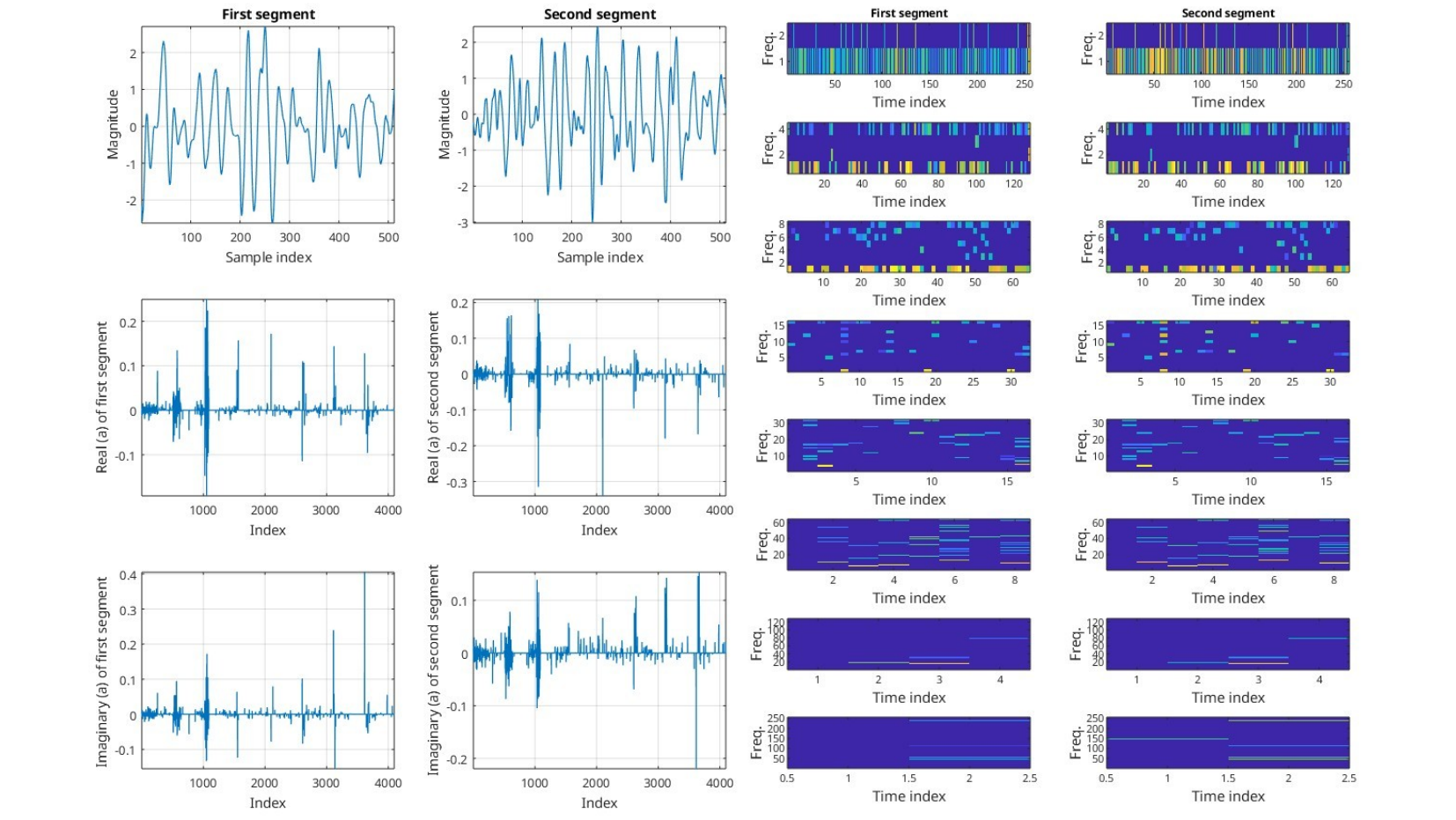}
\caption{Feature extraction from two murmur segments from the same heart sound recording of diamond type decomposed together using COMP and the developed multiresolution complex Gabor dictionary. Columns from left to right: the first murmur segment and the real and imaginary of its decomposition, $\mathbf{a}$ followed by the second murmur segment and the real and imaginary of its decomposition $\mathbf{a}$, and the reshaped time--frequency representations $\mathbf{A}_j$ of the absolute of the corresponding $\mathbf{a}_j$ for each segment.}
\label{fig:segment2}
\end{figure*}

The feature extraction module is developed to capture the intricate time--frequency characteristics of heart murmurs, as shown in Figure \ref{fig:module}. The module employs a multiresolution dictionary $\mathbf{D}$ composed of complex Gabor atoms that enables the representation of murmur signals across varying time--frequency resolutions. To achieve this, murmur segments are projected onto the dictionary, allowing the extraction of detailed signal features. The projection weights $\mathbf{a}$ are computed using complex orthogonal matching pursuit (COMP), which is a sophisticated algorithm optimized for sparse signal representation. The projection weights are divided into groups, each corresponding to a different resolution. These groups are then reshaped individually to construct 2D time--frequency feature matrices $\mathbf{A}$ that capture various temporal and spectral information scales. This approach ensures that the extracted features effectively capture the temporal and spectral variations in heart murmur signals. By providing multiple resolutions of time--frequency features, the module enhances the capability of the system to discern subtle differences between different types of heart murmurs. Figure \ref{fig:segment} illustrates how a murmur segment is transformed into multiple time--frequency representations with varying resolutions. Figure \ref{fig:segment2} demonstrates that two segments from the same recording exhibit nearly identical support in the dictionary, although their complex weights differ in phase and magnitude. This enables the reshaped matrices to share identical spectrotemporal atom locations but with adaptive intensity and phase modulation. Such consistency is unattainable if phase is discarded, as the atoms selected per segment become decoupled.
\subsubsection{Complex Gabor dictionary} 
Unlike Fourier bases, Gabor atoms overlap, providing a flexible and adaptive representation that aligns better with the structures of real-world signals. This flexibility stems from the construction of Gabor atoms using a Gaussian window function, which is known for optimal time--frequency localization. The width of the Gaussian window can be scaled to allow fine-tuning of the resolution in the time and frequency domains. Because the window is correlated with the signal, the Gabor atoms capture localized features, making them effective for signals with transient or nonstationary properties. 

In addition, modulating the Gaussian window generates oscillatory components, allowing the representation of quasiperiodic patterns in the signal. This combination of scaling, translation, and modulation creates a redundant but highly descriptive dictionary of atoms that can provide a compact and precise decomposition of complex signals. This adaptability makes Gabor transforms and related methods particularly valuable in applications such as speech processing, image analysis, and biomedical signal interpretation, where the ability to capture both temporal and spectral characteristics is crucial. {Heart murmurs differ not only in amplitude envelopes, but also in micro-oscillation patterns and localized phase rotations caused by turbulent flow. Complex Gabor atoms encode these phase variations, which real-valued atoms cannot recover}. The complex Gabor atom is defined as, per \cite{mah2024},
\begin{figure}[t!]
\centering
\includegraphics[scale=0.90,trim={0.5cm 0cm 0cm 0cm}]{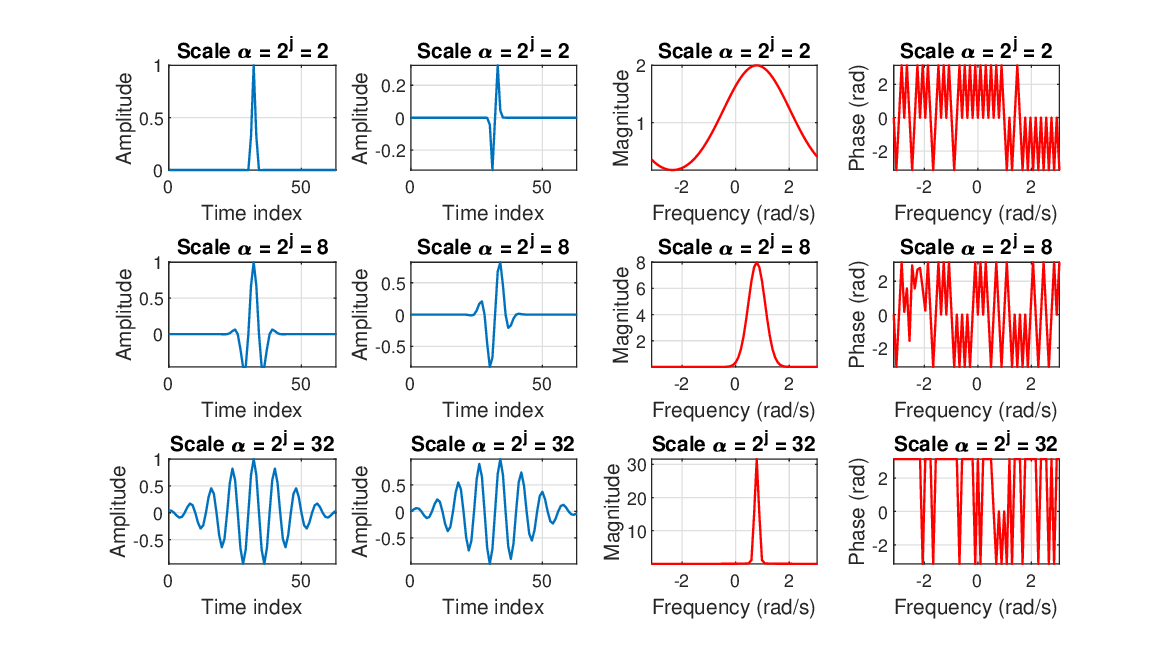}
\caption{Examples of complex atoms with three different resolutions for $\alpha=2^j$ values, with $j=1, 3, 5$. The first two columns on the left show the real and imaginary parts of the atoms in the time domain, respectively. The third and fourth columns show the magnitudes and phases of the corresponding frequency representations, respectively. Plots are for $M=64$, $\omega=\pi/2$, and $m_0=32$. Higher $j$ increases spectral but reduces temporal resolution.}
\label{fig:atom}
\end{figure}
\begin{equation}
d_{(\alpha,m_0,\omega)}(m)= e^{-\pi (\frac{m-m_0}{\alpha})^2} e^{-i(\omega (m-m_0))},
\label{eq:atom}
\end{equation}
where $i=\sqrt{-1}$, and $m=0,1,2,..., M-1$. The Gaussian window limits the time extent of the complex Gabor function, thereby allowing localized analysis. To maintain the overlap among the Gabor atoms at $50\%$, the scale parameter $\alpha$ was first chosen as a power of $2$, $\alpha  = 2^j$ with $0<j<\log_2(M)$. The translation $m_0$ and frequency $\omega$ parameters are then obtained as follows \cite{Qiu1995DiscreteGS}:
\begin{equation}
m_0 = 2^j (0:1: 2^{-j}M-1)~\text{and}~\omega = 2^{-j} (0:1: 2^j-1) 2\pi.
\end{equation}
The total number of frequency bins $\omega$ is equal to $2^{j}$ for each translation step $m_0$ for the total number of steps $2^{-j}M$. The number of atoms is $2^{-j}M \times 2^{j} = M$. The $j$th Gabor dictionary is built for a specific value of $j$ by arranging the column atoms side-by-side in a matrix, that is, $\mathbf{D}_j=[\mathbf{d}_1\mathbf{d}_2...\mathbf{d}_m]\in \mathbb{C}^{M \times M}$. Figure \ref{fig:atom} shows plots of three examples of complex atoms with their corresponding frequency representations.

{The mathematical derivation of the Gabor dictionary parameters ($\alpha$, $\omega$, $m_0$), the justification of the $50\%$ overlap criterion, and the validation of COMP sparsity levels have been comprehensively presented in our previously published work \cite{mah2024}. To avoid redundancy, we refer readers to that publication for the complete theoretical and experimental validation underlying the dictionary construction.}

\subsubsection{Multiresolution complex Gabor dictionary} 
In the above representation of the atoms, the resolution of the dictionary is intricately governed by parameter \( j \), which determines the balance between the temporal and spectral resolutions. Specifically, as \( j \) increases, the real exponential function associated with the atom exhibits a slower decay around the central time \( m_0 \). This slower decay causes the duration of the atom to extend, resulting in a significant degradation of the temporal resolution. In other words, the atom becomes less precise when isolating transient events, leading to a smeared effect in the time domain. Moreover, the extended duration of an atom has a complementary effect on its spectrum. As the temporal span of the atom increases, its bandwidth decreases correspondingly, causing the frequency representation of the atom to become more narrowly focused around its center frequency \( \omega \). This bandwidth narrowing enhances the spectral resolution, making the dictionary more adept at capturing fine-grained details in the frequency domain. Consequently, the trade-off governed by \( j \) highlights the limitation of time--frequency representations, where improving the resolution in one domain inevitably comes at the expense of the resolution in the other. By carefully selecting \( j \), the dictionary can be tailored to prioritize the temporal or spectral resolution, depending on the specific requirements of the signal analysis, underscoring the critical role of this parameter in constructing effective representations.

Because the time--frequency resolution varies with the variable \( j \), the multiresolution Gabor dictionary $\mathbf{D} \in \mathbb{C}^{M \times {JM}}$ is formed by arranging the single-resolution dictionaries $\mathbf{D}_j \in \mathbb{C}^{M \times M}$ for \( j = 1, \ldots, J \) next to each other in a large matrix such as
\begin{equation}
\mathbf{D}=
\begin{bmatrix}
         \mathbf{D}_1|& \mathbf{D}_2| & \cdots & |\mathbf{D}_J
     \end{bmatrix}
     ,~\text{with}~ J={\log_2(M)-1}.
\end{equation}

\subsubsection{Complex orthogonal matching pursuit (COMP)} 
The orthogonal matching pursuit (OMP) is a widely used greedy algorithm designed to obtain sparse solutions for linear systems by leveraging measurements and a predefined dictionary of potential basis elements. OMP operates iteratively by selecting the best-matching atoms, individual components, or dictionary columns that align most closely with the residual signal at each step. Starting with the original signal as the initial residual, the algorithm identifies the atom that maximizes its absolute correlation with the current residual, thereby capturing the most significant signal component in a given dictionary. Once an atom is selected, the OMP updates the approximation of the signal by solving a least-squares problem over the set of chosen atoms, ensuring that the resulting representation is optimal in terms of minimizing errors. The residual is then recomputed as the difference between the original signal and the current approximation, and the process is repeated until a specified stopping criterion is satisfied, such as a predefined sparsity level or an error threshold. This iterative selection and refinement process enables the OMP to construct a sparse approximation of the signal, making it particularly useful in applications such as compressed sensing, image processing, and feature selection, where sparsity is a desirable property for efficient representation and analysis.

Similar to the OMP, the complex orthogonal matching pursuit (COMP) finds a complex sparse vector $\mathbf{a} \in \mathbb{C}^{JM}$ given a measurement vector $\mathbf{x} \in \mathbb{R}^M$ and a dictionary $\mathbf{D} \in \mathbb{C}^{M \times {JM}}$, such that \cite{Rish2015SparseMT,Pra2021FastMP}
\begin{equation}
\mathbf{x}\approx \mathbf{D}\mathbf{a}.
\end{equation}

The best $\zeta$-approximation vector $\mathbf{a}$ of the signal $\mathbf{x}$ is obtained by solving the following problem:
\begin{equation}
\arg\min_{\mathbf{a}}||\mathbf{x}-\mathbf{D}\mathbf{a}||^2_2 ~~ \text{subject to}   ~~ ||\mathbf{a}||_0\leq \zeta,
\label{eq:sparse}
\end{equation}
where $||\mathbf{x}-\mathbf{D}\mathbf{a}||^2_2$ is the least-squares term that measures the reconstruction error between the original signal $\mathbf{x}$ and its approximation $\mathbf{D}\mathbf{a}$, and $||\mathbf{a}||_0$ is the $l_0$-norm that promotes sparsity in $\mathbf{a}$. The problem in \eqref{eq:sparse} can be solved using greedy methods, such as COMP. Recall that the main goal is to find the best $\zeta$-term approximation of a given signal $\mathbf{x}$ by elements from the multiresolution complex Gabor dictionary $\mathbf{D}$ so that
\begin{equation}
\mathbf{x}\approx \mathbf{D}\mathbf{a}=
\begin{bmatrix}
\mathbf{D}_1| & \mathbf{D}_2| & \cdots & |\mathbf{D}_J
\end{bmatrix}
\begin{bmatrix}
\mathbf{a}_1\\
\mathbf{a}_2\\
\vdots\\
\mathbf{a}_J
\end{bmatrix}.
\end{equation}

The entries of vector $\mathbf{a}_j$ describe the contribution of atoms in the $j$th dictionary $\mathbf{D}_j$. For multiple murmur segments, the main goal is to find complex sparse multiple vectors $\mathbf{b}^1\in\mathbb{C}^{JM}$, $\mathbf{b}^2\in\mathbb{C}^{JM}$, ..., $\mathbf{b}^U\in\mathbb{C}^{JM}$ given multiple signals $\mathbf{x}^1\in \mathbb{R}^M$, $\mathbf{x}^2\in\mathbb{R}^M$, ..., $\mathbf{x}^U\in\mathbb{R}^M$ using the shared Gabor dictionary $\mathbf{D}$, so that
\begin{equation}
[\mathbf{x}^1, \mathbf{x}^2, ..., \mathbf{x}^U] \approx \mathbf{D}[\mathbf{b}^1, \mathbf{b}^2, ..., \mathbf{b}^U]=
\begin{bmatrix}
\mathbf{D}_1| & \mathbf{D}_2| & \cdots & |\mathbf{D}_J
\end{bmatrix}
\begin{bmatrix}
\begin{bmatrix}
\mathbf{b}^1_1\\
\mathbf{b}^1_2\\
\vdots\\
\mathbf{b}^1_J
\end{bmatrix}
\begin{bmatrix}
\mathbf{b}^1_1\\
\mathbf{b}^1_2\\
\vdots\\
\mathbf{b}^1_J
\end{bmatrix}, ...,
\begin{bmatrix}
\mathbf{b}^U_1\\
\mathbf{b}^U_2\\
\vdots\\
\mathbf{b}^U_J
\end{bmatrix}
\end{bmatrix}.
\end{equation}

The entries of vector $\mathbf{b}^u_j$ define the contribution of atoms in the $j$ dictionary $\mathbf{D}_j$ to model the signal $\mathbf{x}^u$. {Variable $U$ denotes the number of segments, which varies depending on the number of segments in one recording.} The COMP algorithm for modeling single and multiple murmur segments based on the multiresolution Gabor dictionary  $\mathbf{D}_j$ is summarized in Algorithm \ref{algo:1}.
\begin{algorithm}[t!]
\caption{Modeling single or multiple murmur segments using COMP}
\begin{algorithmic}[1]
\State \textbf{Input:} single segment $\mathbf{x} \in \mathbb{R}^M$ or multiple segments $\mathbf{x}^1\in \mathbb{R}^M$, $\mathbf{x}^2\in\mathbb{R}^M$, ..., $\mathbf{x}^U\in\mathbb{R}^M$, dictionary $\mathbf{D} \in \mathbb{C}^{M \times {JM}}$, sparsity level $\zeta$
\State \textbf{Initialize:} residual $\mathbf{r}_0 = \mathbf{x}$ or 
$\mathbf{r}^1_0=\mathbf{x}^1$, $\mathbf{r}^2_0=\mathbf{x}^2$, ..., $\mathbf{r}^U_0=\mathbf{x}^U$, $\mathbf{a} \in \mathbf{0}^{JM}$ or $\mathbf{b}^1\in\mathbf{0}^{JM}$, $\mathbf{b}^2\in\mathbf{0}^{JM}$, ..., $\mathbf{b}^U\in\mathbf{0}^{JM}$, index set $\Lambda = \emptyset$, counter $c = 0$.
\While{$c < \zeta$}
\State $c = c + 1$
\State $\mathbf{g}=|\mathbf{D}^H \mathbf{r}_{c-1}|$ or $=\sum_u|\mathbf{D}^H [\mathbf{r}^1_{c-1}, \mathbf{r}^2_{c-1}, ..., \mathbf{r}^U_{c-1}]|$ \Comment{$.^H$ is the hermitian conjugate}
\State $i_c = \arg\max_{i} \mathbf{g}_i$ \Comment{Find the index of a matched atom}
\State $\Lambda = \Lambda \cup \{i_c\}$ \Comment{Update the index set of matched atoms}
\State $\mathbf{D}_\Lambda = [\mathbf{d}_{i_1}, \mathbf{d}_{i_2}, \ldots, \mathbf{d}_{i_c}]$ \Comment{Form the sub-dictionary of matched atoms}
\State $\mathbf{a}_\Lambda = \mathbf{D}_\Lambda^{\dagger} \mathbf{x}$ or $[\mathbf{b}^1_\Lambda, \mathbf{b}^2_\Lambda, ..., \mathbf{b}^U_\Lambda] =\mathbf{D}_\Lambda^{\dagger} [\mathbf{x}^1, \mathbf{x}^2, ..., \mathbf{x}^U]$ \Comment{$.^{\dagger}$ is the pseudo inverse}
\State $\mathbf{r}_c = \mathbf{x} - \mathbf{D}_\Lambda \mathbf{a}_\Lambda$ or $[\mathbf{r}^1_{c}, \mathbf{r}^2_{c}, ..., \mathbf{r}^U_{c}] = [\mathbf{x}^1, \mathbf{x}^2, ..., \mathbf{x}^U] - \mathbf{D}_\Lambda [\mathbf{b}^1_\Lambda, \mathbf{b}^2_\Lambda, ..., \mathbf{b}^U_\Lambda]$\Comment{Update}
\EndWhile
\State \textbf{Output:} $\mathbf{a}$ or $[\mathbf{b}^1, \mathbf{b}^2, ..., \mathbf{b}^U]$ are with complex entries, $\mathbf{a}_\Lambda$ or $[\mathbf{b}^1_\Lambda, \mathbf{b}^2_\Lambda, ..., \mathbf{b}^U_\Lambda]$, respectively, at indices defined in $\Lambda$.
\end{algorithmic}
\label{algo:1}
\end{algorithm}

Although OMP can be used with real atoms, it discards the Hermitian symmetry and phase of analytic complex Gabor atoms. Phase encodes instantaneous frequency drift and timing of peak turbulence, which are key discriminants for crescendo vs. diamond murmurs (Figure \ref{fig:mur}). The complex projection of COMP preserves this information in the weights $\mathbf{a} \in \mathbb{C}^{JM}$, which is leveraged implicitly in $||\mathbf{A}_j||^2$ and explicitly in the coherence modeling (Figure \ref{fig:segment2}). COMP extends OMP to complex-valued atoms, preserving phase coherence, which is also critical for joint modeling of multiple segments. 

Unlike conventional greedy algorithms such as fast matching pursuit (FastMP) \cite{Pra2021FastMP}, compressive sampling matching pursuit (CoSaMP) \cite{Zhang2016ComparisonOF}, or subspace pursuit (SP) \cite{Dai2008SubspacePF}, COMP supports joint sparse coding with a shared support $\Lambda$, enabling consistent time--frequency atom selection across variable murmur segments. This shared support is not achievable with real OMP or greedy alternatives without ad-hoc, post-hoc alignment. Moreover, FastMP, CoSaMP, and SP assume uniform, nonoverlapping supports, which are incompatible with the highly redundant multiresolution dictionary. Their iterative pruning mechanisms remove atoms from early resolutions, which degrades feature interpretability.

\subsubsection{Reshaping vectors into matrices}
To construct a time–frequency representation $A_j$, each coefficient vector $a_j$ is reshaped into a matrix whose dimensions preserve the inherent structure of the atoms in the $j$th Gabor dictionary $D_j$. As defined in Eq (4.2), the dictionary at scale $j$ contains $2^{j}$ frequency bins and $2^{8-j+1}$ translation steps, resulting in a total of $M = 512$ atoms. The complex coefficient vector $a_j \in \mathbb{C}^{512}$ produced by COMP thus naturally corresponds to a two-dimensional grid indexed by frequency (rows) and translation (columns) of the image. Reshaping $a_j$ into a matrix $A_j \in \mathbb{R}^{2^{j} \times 2^{8-j+1}}$ preserves this intrinsic organization.

This aspect ratio reflects the resolution trade-off in the multiresolution dictionary: lower values of $j$ provide high temporal and low spectral resolution, whereas higher values of $j$ yield low temporal and high spectral resolution. By maintaining this structure, the reshaped matrices remain consistent with the scale-dependent time–frequency characteristics of systolic murmurs, capturing both the spectral concentration and temporal evolution typical of different murmur patterns. Figures \ref{fig:segment} and \ref{fig:segment2} illustrate examples of the resulting representations for $M = 512$ and $J = 8$.

\subsection{Feature classification model}
\begin{figure}[!]
\centering
\includegraphics[width=\textwidth,trim={0cm 0cm 0cm 0cm}]{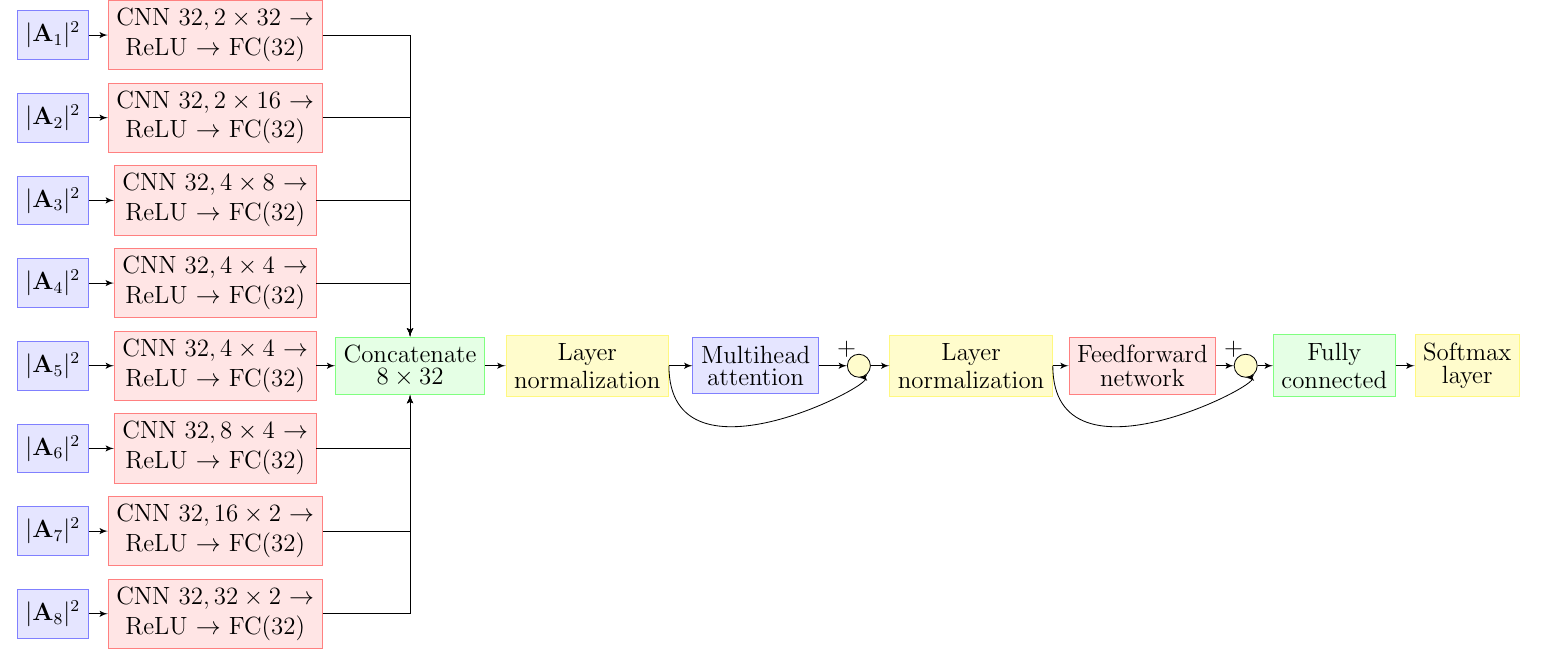}
\caption{Block diagram of the developed vision transformer encoder-based model with multiple convolutional neural networks and one encoder layer.}
\label{fig:model} 
\end{figure}

The model is constructed using a transformer encoder by processing multiple time--frequency feature matrices of varying sizes, as shown in Figure \ref{fig:model}. A feature matrix is obtained by calculating the squared magnitude of the entries of $|\mathbf{A}_j|^2$. Each matrix is then fed into a dedicated network branch comprising a two-dimensional convolutional neural network and a fully connected layer to reduce the dimensionality. The outputs from all branches are concatenated and normalized to form an embedding containing relevant information on systolic heart murmurs. The embedding computation process is detailed in Subsection \ref{subsec:embedding}. Subsequently, the embedding is passed to a multihead attention block to effectively model the interdependencies between all feature representations, as explained in Subsections \ref{subsec:self-attention} and \ref{subsec:multihead}. The output of the multihead attention is then normalized using a normalization layer to stabilize and accelerate training. The normalized output is forwarded to a feedforward network (FFN) consisting of a CNN with a kernel size of one, which enables channel-wise transformations while preserving the spatial dimensions, as explained in Subsection \ref{subsec:feedforward}. Both the multihead attention and FFN are equipped with residual connections, as described in Subsection \ref{subsec:residual_connections}, which aim to improve the gradient flow. After processing through the transformer, the output is forwarded to a fully connected layer, followed by a softmax layer for final classification. This architectural design ensures that the model effectively integrates and processes diverse time--frequency feature matrices while leveraging the strengths of both convolutional operations and transformer mechanisms for a robust performance.

The transformer architecture diverges from traditional recurrent networks, enabling parallel computation and achieving the highest performance for various tasks compared with other architectures. The transformer encoder, which is a critical component of this architecture, excels at modeling long-range dependencies within sequential data by leveraging self-attention mechanisms. This capability is indispensable for language modeling, time series analysis, and speech processing. The transformer encoder is a stack of identical layers, each comprising two main components: a multihead self-attention mechanism and a position-wise feedforward neural network. These components are augmented by residual connections and layer normalization to ensure stability and efficiency during training.

\subsubsection{Patch embedding and tokenization}
\label{subsec:embedding}
Patch embedding and tokenization processes involve dividing an input image into fixed-size patches, which are then tokenized into input tokens. Each patch is projected into a high-dimensional embedding space using a learnable linear projection implemented via 2D convolutional neural networks (CNNs), which enables compatibility with the transformer architectures. The output embedding for each patch is subsequently transformed into a 1D vector through a fully connected layer, where the output dimensionality corresponds to the number of filters used in the CNNs. All resulting token vectors are then concatenated to form a 2D matrix, where each row represents a token in a batch of length corresponding to the number of CNN filters. This matrix serves as an input representation for the transformer encoder. To enhance training stability, accelerate convergence, and reduce internal covariate shifts, a normalization layer is applied to the output of the embedding and tokenization stages.
\subsubsection{Self-attention mechanism}
\label{subsec:self-attention}
The self-attention mechanism is a key feature of the transformer. Unlike recurrent networks, which process sequential inputs step by step, the self-attention mechanism simultaneously computes the attention scores for all input tokens. This enables the model to focus on the relevant parts of the input sequence regardless of their distance. For each token in the input sequence, the self-attention mechanism generates three vectors: query ($\mathbf{q}$), key ($\mathbf{k}$), and value ($\mathbf{v}$) vectors. These vectors are computed by applying learned weight matrices to the input embeddings. Given an input sequence represented as a matrix $\mathbf{G}$ (where each row is an input token embedding), the transformations are given by
\begin{equation}
\mathbf{Q} = \mathbf{G}\mathbf{W}_Q,~ \mathbf{K} = \mathbf{G}\mathbf{W}_K,~ \text{and}~\mathbf{V} = \mathbf{G}\mathbf{W}_V,     \end{equation}
where $\mathbf{W}_Q$, $\mathbf{W}_K$, and $\mathbf{W}_V$ are learnable weight matrices with dimensionality or number of channels equal to \( d_k \). $\mathbf{Q}$, $\mathbf{K}$, and $\mathbf{V}$ are the matrices of query, key, and value for all input tokens, respectively. The transformed vectors are used to calculate attention scores by computing the scaled dot product between the query and key vectors. These scores are then normalized using a softmax function and applied to the value vectors, resulting in a weighted sum that captures the contextual representation of each token in the input sequence, as follows
\begin{figure}[t]
\centering
\includegraphics[width=\textwidth,trim={0cm 0cm 0cm 0cm}]{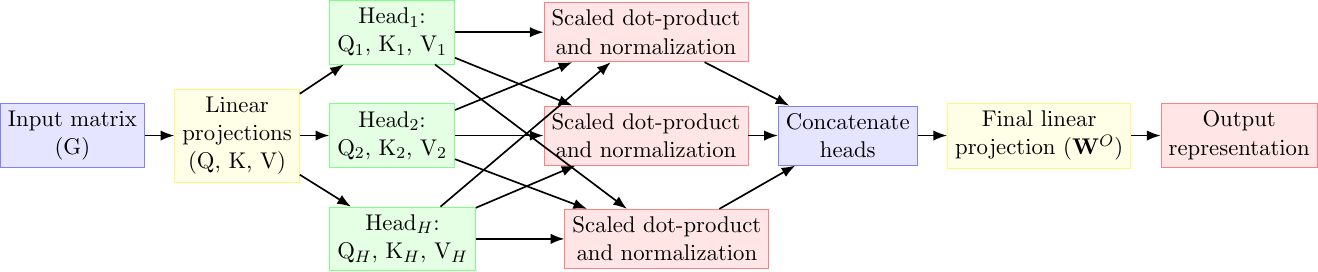}
\caption{Block diagram of the multihead attention with $H$ the total number of heads.}
\label{fig:multihead} 
\end{figure} 
\begin{equation}
\text{Attention}(\mathbf{Q}, \mathbf{K}, \mathbf{V}) = \text{softmax}\left(\frac{\mathbf{Q}\cdot\mathbf{K}^T}{\sqrt{d_k}}\right)\mathbf{V}.
\end{equation}
\subsubsection{Multihead attention}\label{subsec:multihead}
The multihead attention mechanism extends the single-head self-attention mechanism to capture various relationship patterns in data. It projects queries, keys, and values onto multiple subspaces (heads) using a learnable linear transformation. Each head independently computes its attention output, which is then concatenated and linearly transformed to obtain the final attention output. This diversity allows the model to address multiple aspects of the input simultaneously, thereby enhancing its representational capabilities. This process is expressed in \eqref{eq:multi_head}. 
\begin{equation}
\begin{split}
\text{multihead}(\mathbf{Q}, \mathbf{K}, \mathbf{V}) = & \text{Concatenate}(\text{head}_1, \ldots, \text{head}_H)\mathbf{W}^O\\ \text{head}_h = & \text{Attention}(\mathbf{Q}\mathbf{W}^h_Q, \mathbf{K}\mathbf{W}^h_K, \mathbf{V}\mathbf{W}^h_V),
\end{split}
\label{eq:multi_head}
\end{equation}
where $\mathbf{W}^h_Q$, $\mathbf{W}^h_K$, and $\mathbf{W}^h_V$ are learnable weight matrices for the corresponding $h$ head. Finally, a linear transformation $\mathbf{W}^O$ combines the heads into the expected output dimension. Figure \ref{fig:multihead} illustrates the multihead attention mechanism.
\subsubsection{Position-wise feedforward neural network}
\label{subsec:feedforward}
Following multihead attention, each token passes through an FFN that operates independently at every position. This FFN comprises a linear transformation using a CNN with a filter kernel of size one, followed by a ReLU, as expressed in \eqref{eq:ffn}. This enhances the capacity of the encoder to learn the complex transformations of the input embeddings.
\subsubsection{Residual connections and layer normalization}
\label{subsec:residual_connections}
To mitigate the vanishing gradient problem and enhance the gradient flow during backpropagation, the transformer encoder incorporates residual connections around both the multihead attention and FFN sublayers. These residual connections help to preserve information and facilitate more effective weight updates. Following each residual connection, layer normalization is applied to the outputs, as indicated in \eqref{eq:residual_connections}. This stabilizes training, accelerates convergence, and reduces internal covariate shifts. This combination of residual connections and layer normalization ensures smoother optimization, allowing the model to learn deeper representations.
\begin{equation}
\text{Output} = \text{LayerNorm}(\mathbf{z} + \text{sublayer}(\mathbf{z})).
\label{eq:residual_connections}
\end{equation}
\begin{equation}
\text{FFN}(\mathbf{y}) = \max(0, \mathbf{y}\mathbf{W} + \mathbf{b}).
\label{eq:ffn}
\end{equation}
\section{Experimental analysis and results}
The developed system was experimentally evaluated to measure its effectiveness in distinguishing between the four types of heart murmurs. The evaluation strategy involved assessing the performance of the system using standard classification metrics such as accuracy, specificity, and F1-score. Moreover, the confusion matrix effectively summarizes the classification performance of the model. The datasets were divided into $80\%$ for training and $20\%$ for testing, preserving the class distribution in both subsets to ensure that the model is trained and evaluated on representative examples of each murmur type. The evaluation results were carefully analyzed to identify strengths, limitations, and areas for improvement, guiding potential refinements to enhance performance.
\subsection{System implementation}
Table \ref{tab:system} presents the parameters governing the extraction of multiresolution time--frequency matrices derived from PCG signals along with the configuration details of the hybrid classification model. These parameters define the setup of the feature extraction module, including the size of the multiresolution complex Gabor dictionary and the dimensions of the resulting time--frequency matrices. It can be observed, for example, that the sparsity level $\zeta$, is assigned a maximum value of $511$, which matches the signal length ($512$) not because the method requires a full decomposition but because a large redundant dictionary is used ($8$ resolutions $\times$ $512$ atoms each), which results in a minimum sparsity of ratio$=(4096-511)/4096=0.875$. Increasing such a value by reducing $\zeta$ results in high sparse models, which may underfit the training because the models are transformed into time–frequency matrices.

In addition, we evaluated several configurations of the model structure and training hyperparameters, and the parameter set reported in Table \ref{tab:system} corresponds to the combination that consistently produced the best performance and stable convergence. These optimal settings ensured effective learning without overfitting. The table outlines the key architectural elements of feature classification, such as the size of the convolutional kernels in the CNNs, which are assigned different values depending on the resolution of the input feature matrix. Moreover, it reports the number of multihead attention values and key channels used to implement the transformer encoder. It also details the essential training parameters, including the optimization algorithm (stochastic gradient descent with momentum (SGDM)) and the loss function (cross-entropy).

\begin{table}[t!]
\centering
\caption{Main parameters of the implemented system.}
\label{tab:system}
\begin{tabular}{@{}llll@{}}
\hline
\multicolumn{2}{c}{\text{Feature extraction module}}&\multicolumn{2}{c}{\text{Feature classification model}}\\
\text{Parameter}&\text{Value}&\text{Parameter}&\text{Value}\\
\hline
\text{\textbf{Murmur:} } & &\textbf{$\text{CNN}$, $j=1,2,...,8$:} &\\
\text{Segment length (samples)} & $512$ & \text{no. kernel filters}&$32$\\
\text{\textbf{Dictionary:}}& &\text{Filter size ($j$)} &
$2^{{j/1.75}} \times 2^{6-{j/1.75}}$\\
\text{Resolution} & $j=1,2,...,8$& \text{Filter stride ($j$)} &
$2^{{j/1.75}} \times 2^{6-{j/1.75}}$\\
\text{Size of Dictionary:} & $512\times (512*8)$&\textbf{Transformer:}&\\
\textbf{COMP:} & & \text{no. heads}&$2,4,8,...$, or $256$\\
\text{no. iterations (sparsity level)}&511& \text{no. key channels}& \text{no. heads}\\
\text{Length of projection vector $\mathbf{a}$} & $512*8$&\textbf{Training:}&\\
\textbf{Matrices $\mathbf{A}_j$, $j=1,2,...,8$:} & &\text{Optimizer}&\text{SGDM}\\
\text{Size of matrix $j$}& $2^j\times 2^{8-j+1}$&\text{Loss function}&\text{crossentropy}\\
 & &\text{Batch size}&$150$\\
 & &\text{no. epochs}&$500$\\
 & &\text{no. classes}&$4$\\
\hline\end{tabular}
\end{table}

\subsection{System computational complexity}
The computational complexity of a neural network refers to the computational resources required for training and inference processes. This complexity depends on factors such as network depth (the number of layers and learnable parameters). Deeper networks involve more computations during forward and backward propagation, and a higher parameter count increases memory usage and training duration. Excessive complexity can hinder convergence and increase overfitting risks, necessitating an efficient architectural design and regularization to balance performance and computational efficiency. Table \ref{tab:complexity} lists the number of trainable parameters of the model in terms of the number of attention heads and dimensions of each head. Increasing the number of attention heads or the dimensionality of each head leads to a greater system complexity.
\subsection{Evaluation metrics}
To evaluate the performance of the proposed multiclass murmur classification system, we employed a set of metrics widely used in multiclass diagnostic applications. For each murmur class, we computed the true positives (TP), false positives (FP), false negatives (FN), and true negatives (TN), from which class-specific performance measures were derived. However, due to the strong imbalance among the four murmur types—particularly the limited number of crescendo recordings—class-wise metrics alone do not provide a sufficiently fair overall assessment. Therefore, in addition to reporting per-class specificity, F1-score, and accuracy, we also include \emph{macro-averaged} metrics, which treat all classes equally regardless of their sample sizes. These metrics are defined as follows:
\begin{equation}
\text{Macro-Specificity} = \frac{1}{C} \sum_{i=1}^{C} 
\frac{TN_i}{TN_i + FP_i},
\end{equation}
\begin{equation}
\text{Macro-F1} = \frac{1}{C} \sum_{i=1}^{C}
\frac{2TP_i}{2TP_i + FP_i + FN_i},
\end{equation}
\begin{equation}
\text{Macro-Accuracy} = \frac{1}{C} \sum_{i=1}^{C}
\frac{TP_i + TN_i}{TP_i + TN_i + FP_i + FN_i},
\end{equation}
where $C=4$ denotes the number of murmur classes. These macro-averaged metrics provide an unbiased evaluation of the performance across all categories, irrespective of class imbalance. To further quantify the stability and robustness of the classifier, all reported metrics are presented as the mean $\pm$ standard deviation obtained from the stratified 5-fold cross-validation. This allows for a reliable comparison across different folds and reduces the impact of sample imbalance on the evaluation.
\begin{table}[t!]
\centering
\caption{The number of trainable parameters in thousands, with "M" denoting a million, for a model of 44 layers, in terms of the number of attention heads and key or query channels.}
\label{tab:complexity}
\begin{tabular}{llllllllllllllll}
\hline
\multirow{2}{*}{\diagbox[width=10em]{Channels($d_k$)}{Heads(H)}}&\text{single-head}&\multicolumn{8}{c}{multihead}\\
&1&2&4&8&16&32&64&128&256\\
\hline
$1$ $\times$ \text{no. heads}&159.5&159.6&159.9&160.4&161.5&163.6&167.8&176.1&192.9\\
$2$ $\times$ \text{no. heads}&159.6&159.9&160.4&161.5&163.6&167.8&176.1&192.9&226.5\\
$4$ $\times$ \text{no. heads}&159.9&160.4&161.5&163.6&167.8&176.1&192.9&226.5&293.5\\
$8$ $\times$ \text{no. heads}&160.4&161.5&163.6&167.8&176.1&192.9&226.5&293.5&427.7\\
$16$ $\times$ \text{no. heads}&161.5&163.6&167.8&176.1&192.9&226.5&293.5&427.7&696\\
$32$ $\times$ \text{no. heads}&163.6&167.8&176.1&192.9&226.5&293.5&427.7&696&1.2 M\\
$64$ $\times$ \text{no. heads}&167.8&176.1&192.9&226.5&293.5&427.7&696&1.2 M&2.3 M\\
$128$ $\times$ \text{no. heads}&176.1&192.9&226.5&293.5&427.7&696&1.2 M&2.3 M&4.4 M\\
\hline
\end{tabular}
\end{table}

\subsection{Training dynamics and convergence analysis}
\begin{figure}[t!]
\centering
\includegraphics[width=1\textwidth,trim={0cm 0cm 0cm 0cm}]{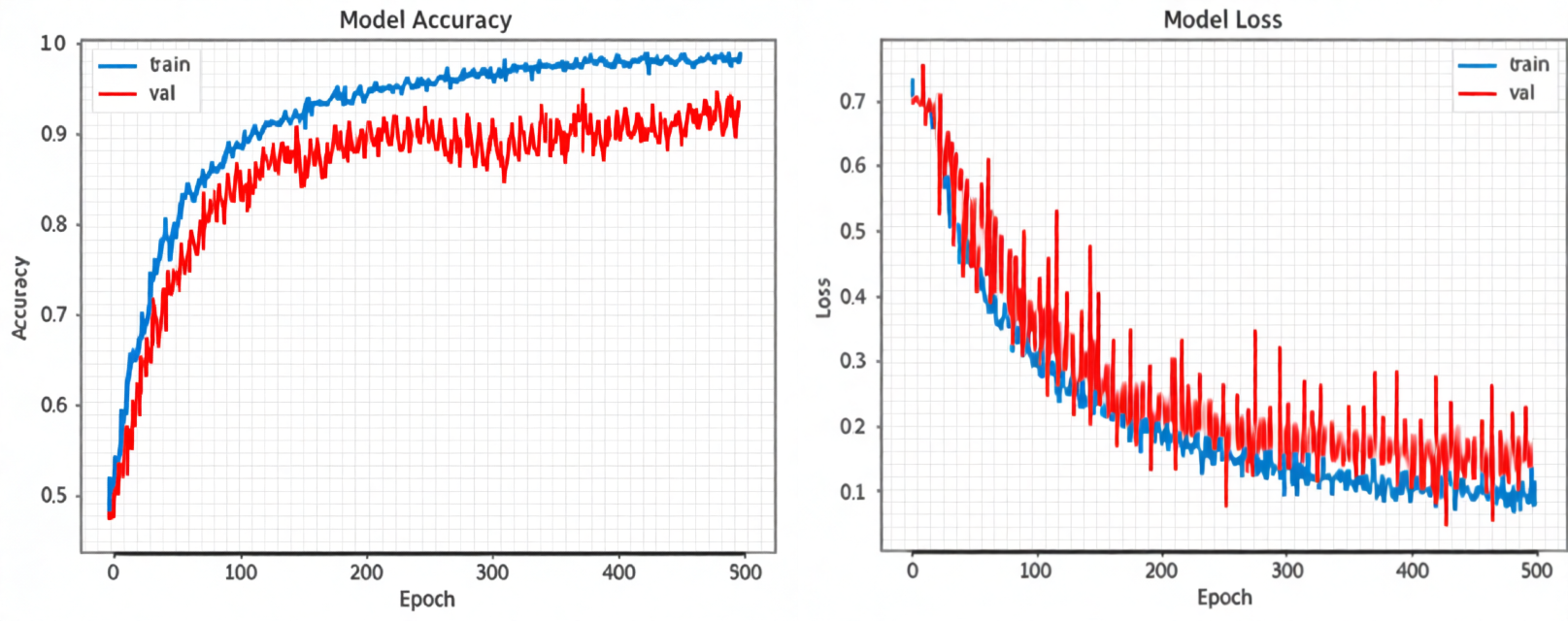}
\caption{The evolution of the training and validation loss and accuracy across epochs.}
\label{fig:train} 
\end{figure}

To examine the learning behavior of the proposed system and assess potential underfitting or overfitting, we monitored the evolution of the training and validation loss and accuracy across epochs. Figure~\ref{fig:train} presents the corresponding curves for the best-performing configuration.

The training and validation accuracy curves exhibit similar increasing trends and converge toward high final values, with no observable divergence between the two. Similarly, the training and validation loss curves decrease smoothly and remain closely aligned throughout the optimization process. This consistent behavior indicates that the model generalizes well and does not overfit the training data, despite the inherent class imbalance and the limited number of crescendo samples.

Furthermore, the absence of oscillations or abrupt fluctuations in the validation curves suggests stable optimization dynamics. This stability is attributed to the combination of the SGDM optimizer and the multistage multiresolution Gabor projection, followed by the transformer encoder, which together promote robust feature learning. These observations confirm that the proposed architecture can effectively learn discriminative representations without suffering from underfitting or overfitting, even under challenging data conditions.

\subsection{Results}
We conducted experiments using $5$-fold cross-validation ($5$-fold CV) to assess system performance. In this process, the extracted feature matrices of murmur segments from each class were divided into five nonoverlapping subsets. The model was then trained and evaluated five times, each time using four subsets for training and the remaining subset for testing. This stratified approach ensures that all matrices are used for both training and validation across different folds, minimizing the risk of overfitting and providing a more reliable estimate of the generalization of the model.

Table \ref{tab:HandD} lists the average classification accuracy of the proposed system as a function of the number of attention heads ($H$) and the dimensionality of the output from each head, denoted by the number of channels ($d_k$) in the transformer encoder. These two parameters control the capacity of the model to capture multiresolution contextual relationships within the multisegment time--frequency representations, and modifying them effectively reflects the impact of simplifying or altering the attention mechanism. 

Multihead attention improves classification accuracy over single-head configurations by allowing the model to capture a more diverse set of discriminative features from input representations. It can also be observed that a trade-off exists between the number of attention heads and the dimensionality of each head. Specifically, configurations with fewer attention heads can still achieve competitive or even improved performance when the dimensionality of each head is increased because the model gains a richer representation capacity in each attention stream. Conversely, a higher number of attention heads can compensate for a lower per-head dimensionality by allowing the model to attend to a wider range of features simultaneously. This complementary relationship suggests that optimal performance does not necessarily require the simultaneous maximization of both parameters; instead, a balanced configuration tailored to the task characteristics can yield superior results. These findings emphasize the flexibility of the transformer architecture and the importance of strategic hyperparameter tuning.
\begin{table*}[t!]
\centering
\caption{The average classification macro-accuracy over 5-fold CV on the test set is reported as a function of the number of attention heads and the dimensionality of each head or the number of channels.}
\label{tab:HandD}
\begin{tabular}{llllllllllllll}
\hline
\multirow{2}{*}{\diagbox[width=10em]{Channels($d_k$)}{Heads(H)}}&\text{single-head}&\multicolumn{8}{c}{multihead}\\
&1&2&4&8&16&32&64&128&256\\
\hline
{$1$ $\times$ \text{no. heads}}&95.46&95.10&94.43&94.61&94.69&94.69&95.39&95.53&95.13\\
{$2$ $\times$ \text{no. heads}}&94.69&95.61&94.54&94.69&95.28&95.39&95.20&95.72&94.87\\
{$4$ $\times$ \text{no. heads}}&94.50&94.95&95.53&95.35&95.28&94.76&94.76&95.61&93.25\\
{$8$ $\times$ \text{no. heads}}&94.50&95.02&95.28&94.95&95.35&95.02&95.98&94.76&94.43\\
{$16$ $\times$ \text{no. heads}}&94.43&94.61&95.46&95.95&95.61&94.28&95.13&92.81&93.58\\
{$32$ $\times$ \text{no. heads}}&94.47&95.72&95.96&94.95&94.03&94.61&94.61&93.33&93.66\\
{$64$ $\times$ \text{no. heads}}&94.43&95.89&94.76&94.95&94.61&95.02&93.85&93.58&92.74\\
{$128$ $\times$ \text{no. heads}}&94.50&94.28&94.50&94.28&91.96&93.77&93.07&93.00&92.22\\
\hline
\end{tabular}
\end{table*}

\subsection{Comparison}
To evaluate the performance of the developed system, we conducted a thorough comparative analysis by implementing and testing several systems based on well-established signal processing techniques and deep learning models. Specifically, we built baseline systems using the short-time Fourier transform (STFT), continuous wavelet transform (CWT), and Wigner–Ville distribution (WVD) \cite{Akay1996}, in conjunction with residual network (ResNet) \cite{He2015DeepRL} and parallel CNNs-transformer (PCTMF-Net) \cite{Wang2023PCTMFNetHS}. 

These systems serve as representative benchmarks for both traditional signal analysis methods and modern deep learning approaches. The STFT extracts spectral features, the CWT captures multiresolution characteristics, and the WVD provides a high-resolution time--frequency representation of the input signals. Input representations were directly used as input to ResNet and PCTMF-Net, whereas they were standardized to $8$ feature matrices (via resampling, tiling, or patching) for the proposed model. The extracted information is fed to a ResNet, which leverages its hierarchical feature learning capability to classify the inputs. The PCTMF-Net can process the extracted information through a two-way parallel CNN–transformer architecture with second-order spectral features. Table \ref{tab:net} lists the settings of the implementation parameters for STFT, CWT, and ResNet.

Table \ref{tab:comp} presents a comparative evaluation of different feature extraction methods and classification models in terms of average classification accuracy, specificity, and F1-score, along with the corresponding standard deviations across the 5-fold CV. The results demonstrate the performance of two main classification architectures, ResNet and the proposed transformer-based model, when applied to different time--frequency representations of the input signals, namely the STFT, CWT, and resampled CWT. {The number of attention heads and dimensionality of each head in the transformer encoder were set to the best-performing values obtained from the deep analysis results reported in Table \ref{tab:HandD}, which corresponded to the case of using four heads with a dimensionality of $32$.}

The highest accuracy of $95.96\%$ is achieved by the second variant of the proposed system, Proposed II (projection of multiple murmurs); it also attains the highest specificity of $98.43\%$, with the lowest standard deviations among all methods. This indicates superior performance in correctly identifying the types of murmur segments, particularly in minimizing false positives. The corresponding F1-score of $81.87\%$ suggests a balanced trade-off between precision and recall for this configuration. These results underscore the effectiveness of projecting multiple murmur segments, which effectively mitigates murmur variability and overlap within a single recording by learning distinct weights that operate over consistent time--frequency distributions.

However, the discrepancy between the high accuracy ($95.96\%$) and the lower macro F1-score ($81.87\%$) for Proposed II arises from the intrinsic class imbalance in the dataset. Although accuracy is dominated by the performance of the majority classes, the macro F1-score more strongly penalizes misclassifications in minority murmur categories. The few errors made by the model occur primarily in these underrepresented classes, leading to a lower F1-score despite overall high classification accuracy. This behavior is consistent with imbalanced multiclass settings and does not indicate instability in the proposed method.

\begin{table*}[t!]
\centering
\caption{\textls[5]{Implementation of STFT, CWT, WVD, ResNet, and PCTMF-Net for comparison purposes.}}
\label{tab:net}
\begin{tabular}{@{}llllll@{}}
\hline
\multicolumn{2}{c}{\text{Features}}&\multicolumn{2}{c}{\text{ResNet}}&\multicolumn{2}{c}{\text{PCTMF-Net}}\\
\text{Parameter}&\text{Value}&\text{Parameter}&\text{Value}&\text{Parameter}&\text{Value}\\
\hline
\textbf{STFT:}& &\text{Initial filter size}&$4\times4$&\multicolumn{2}{c}{\text{Two-way CNN-transfomr}}\\
\text{Window length (samples)}&32&\text{No. initial filter}&$32$&\text{No. of CNNs}&$3$\\
\text{Window overlap (samples)}&16&\text{Initial stride}&$2\times2$&\text{Initial filter size}&$3\times3$\\
\text{STFT size (samples)}&$17\times31$&\text{Stack depth}&$[8, 4, 2]$&\text{Initial stride}&$1\times1$\\
\textbf{CWT:}& &\text{No. filter}&$[16, 32, 64]$&\text{No. of heads}&$4$\\
\text{Wavelet name}& \text{Morse}&&\\
\text{CWT size (samples)}&$61\times512$&&\\
\textbf{WVD:}& &&\\
\text{WVD size (samples)}&$64\times512$&&\\
\hline\end{tabular}
\end{table*}

As shown in Table~\ref{tab:comp}, the proposed transformer-based models consistently outperform both the ResNet and PCTMF-Net baselines across all feature representations. This indicates not only superior discriminative power but also enhanced stability across the cross-validation folds. In contrast, even the strongest baseline (CWT + ResNet) lags by $\sim$$10\%$ in accuracy and $\sim$$6$ points in the F1-score. The progressive gains from Proposed I (single-segment projection) to Proposed II (multisegment projection) confirm that enforcing consistent atom support across segments, thereby reducing intra-recording variability, decisively improves performance. These results validate that the synergy of multiresolution sparse modeling and global attention-based classification yields a more robust and clinically reliable framework for murmur shape identification  than other methods. These results suggest that the integration of advanced feature projection techniques within a transformer-based architecture enhances the classification performance, offering a promising approach for the accurate and reliable diagnosis of heart disease.

Table~\ref{tab:per_class_metrics} reports the per-class specificity and F1-score (mean (standard deviation) over $5$-fold cross-validation) for the best-performing configuration (\textit{Proposed II}). The results confirm high model reliability for the two majority classes, diamond and plateau, with specificity above $98\%$ and F1-scores exceeding $93\%$. For the less frequent decrescendo class, the performance remains robust (specificity = $97.40\%$, F1 = $77.30\%$), albeit with higher variability. Critically, even for the severely underrepresented crescendo class (only $2$ recordings, $27$ segments), the model achieves a specificity of $95.80\%$ and an F1-score of $59.10\%$, indicating nontrivial discriminative capability despite the extreme data imbalance---thus refuting concerns of complete misclassification (e.g., zero recall) and lending clinical credibility to the overall high accuracy.

\begin{table}[t!]
\centering
\caption{Comparison results: average classification accuracy and standard deviation.}
\label{tab:comp}
\begin{tabular}{@{}lllll@{}}
\hline
Features & Classifier & Specificity & F1-score&Accuracy\\
\hline
STFT &          ResNet                &88.78 (0.82)&70.79 (2.03)&81.26 (1.19)\\
STFT &          PCTMF-Net             &89.75 (1.10)&71.93 (2.56)&83.31 (1.36)\\
CWT  &          ResNet                &91.16 (0.74)&75.39 (1.64)&85.92 (1.16)\\
WVD  &          ResNet                & 89.31  (0.93) & 73.05  (2.11) & 83.64  (1.27) \\
CWT  &          PCTMF-Net             & 92.17  (0.85) & 77.03  (2.07) & 87.42  (1.04) \\
STFT & Proposed transformer           &90.26 (1.34)&73.20 (3.38)&86.42 (1.06)\\
CWT  & Proposed transformer           &93.00 (1.34)&77.91 (2.84)&88.90 (0.91)\\
WVD  & Proposed transformer           &93.85 (0.999)&79.08 (2.08)&89.85 (1.60)\\
\multicolumn{2}{l}{{Proposed I (Projection of single murmur)}} &95.15 (0.86)&79.66 (1.89)&91.84 (0.97)\\
\multicolumn{2}{l}{{Proposed II (Projection of multiple murmurs)}}&98.43 (0.62)&81.87 (1.41)&95.96 (0.82)\\
\hline
\end{tabular}
\end{table}
\begin{table}[t!]
\centering
\caption{Per-class classification performance (mean(standard deviation)) for the best-performing configuration (Proposed II, 5-fold CV).}
\label{tab:per_class_metrics}
\begin{tabular}{llll}
\toprule
Class & Specificity & F1-score \\
\midrule
Diamond      & 98.10(1.10) & 93.40(1.90) \\
Plateau      & 99.20(0.50) & 96.90(1.10) \\
Decrescendo  & 97.40(1.50) & 77.30(4.90) \\
Crescendo    & 95.80(2.30) & 59.10(10.60) \\
\bottomrule
\end{tabular}
\end{table}
\section{Conclusions and future work}
This study presents a novel system for the automatic classification of systolic heart murmurs. Using sparse modeling through complex orthogonal matching pursuit, the system projects single or multiple murmur segments onto a shared multiresolution Gabor dictionary to effectively capture the time--frequency features while mitigating the variability and overlap between murmur segments within the same recording. These features were processed using a transformer encoder augmented with convolutional networks for tokenization and multiresolution integration. This model learns hierarchical representations from variable-resolution time--frequency matrices, thereby improving generalization and diagnostic accuracy. 

Evaluated on the CirCor DigiScope dataset, the proposed system achieved high performance, outperforming conventional approaches, such as ResNet-based models, using standard time--frequency transforms, such as STFT and CWT. The results confirm that integrating multiresolution signal projection with transformer-based architectures significantly enhances the performance of medical audio classification tasks. This study demonstrates the potential of combining sparse representation techniques with deep learning models for the automated diagnosis of heart problems.

Future work will focus on extending this research to several key areas. Validating the system on larger, more diverse datasets, including multicenter and other real-world clinical recordings, will help assess its generalizability. Exploring its application to other medical audio tasks, such as the detection of diastolic murmurs or respiratory pathologies, could broaden its impact. Further architectural improvements, including optimized tokenization, specialized attention mechanisms, and enhanced interpretability, will improve performance and clinical trust. Finally, evaluating the system for real-time or point-of-care deployment will be explored.

{Future work will also prioritize collaboration with clinicians to manually annotate diastolic murmur morphology (e.g., early vs. mid-diastolic, rumble vs. decrescendo) in existing datasets or prospective collections. This will enable supervised modeling of diastolic patterns and evaluation of cross-timing generalization (e.g., shared dictionary atoms for systolic vs. diastolic turbulence).}

\section*{Use of AI tools declaration}
The authors declare they have not used Artificial Intelligence (AI) tools in the creation of this article.

\section*{Conflict of interest}
The authors declare that there are no conflicts of interest.

\end{document}